\pdfoutput=1
\documentclass[11pt]{article}

\usepackage[final]{mtsummit25}
\usepackage{copyright}
\usepackage{times}
\usepackage{latexsym}

\usepackage[T1]{fontenc}
\usepackage[utf8]{inputenc}

\usepackage{microtype}

\usepackage{inconsolata}

\usepackage{graphicx}

\title{Introducing Quality Estimation to Machine Translation Post-editing Workflow: An Empirical Study on Its Usefulness}

\author{Siqi Liu \\
  The Hong Kong Polytechnic University\\
  \texttt{si-qi.liu@connect.polyu.hk} \\\And
  Guangrong Dai \\
  Guangdong University of Foreign Studies\\
  \texttt{carldy@163.com} \\\AND
  Dechao Li\thanks{~Corresponding author}  \\
  The Hong Kong Polytechnic University \\
  \texttt{dechao.li@polyu.edu.hk} \\
  }

\begin{document}
\maketitle
\begin{abstract}
This preliminary study investigates the usefulness of sentence-level Quality Estimation (QE) in English-Chinese Machine Translation Post-Editing (MTPE), focusing on its impact on post-editing speed and student translators' perceptions. It also explores the interaction effects between QE and MT quality, as well as between QE and translation expertise. The findings reveal that QE significantly reduces post-editing time. The examined interaction effects were not significant, suggesting that QE consistently improves MTPE efficiency across medium- and high-quality MT outputs and among student translators with varying levels of expertise. In addition to indicating potentially problematic segments, QE serves multiple functions in MTPE, such as validating translators’ evaluations of MT quality and enabling them to double-check translation outputs. However, interview data suggest that inaccurate QE may hinder post-editing processes. This research provides new insights into the strengths and limitations of QE, facilitating its more effective integration into MTPE workflows to enhance translators’ productivity. 
\end{abstract}

\section{Introduction}
In a typical machine translation post-editing (MTPE) workflow, translators still need to spend a certain amount of time and effort on evaluating the quality of machine translation (MT) outputs to determine the cost-effectiveness of MTPE. To be more specific, if the MT output is of acceptable quality, post-editing is feasible; otherwise, translating from scratch may be more efficient. However, this process can be time-consuming, especially when the MT outputs are ultimately deemed unsuitable for post-editing. In order to achieve a quick turnaround, it is therefore necessary to speed up or even automate the process of evaluating whether MTPE is worthwhile \citep{ALVAREZVIDAL2023100125}. 

The cost-effectiveness of MTPE can be evaluated from two different but interrelated perspectives: predicting the MT quality \citep{informatics8030061,specia2010machine}, to see whether translators are going to work with “good” MT or “bad” MT, or predicting MTPE effort \citep{10.3389/fpsyg.2017.01282,liu2024}, to see how much effort, such as post-editing time and editing distance, is required by the PE task. In the field of computer science, both approaches are considered as Quality Estimation (QE)\footnote{However, since PE effort is a complex, multidimensional concept influenced by various factors — including but not limited to MT quality — and is not necessarily linearly related to MT quality \citep{ALVAREZVIDAL2023100125,krings2001repairing}, we argue for a clear distinction between the tasks of predicting MT quality and those of predicting PE effort, rather than grouping them into the same category.}. In contrast to traditional reference-based metrics such as BLEU \citep{papineni-etal-2002-bleu}, QE estimates MT quality \textit{without requiring reference translation} \citep{specia2010machine}, making it particularly relevant for real-world translation scenarios. In the current research, we focus on the QE method that provides MT quality scores, rather than the one that estimates MTPE effort. The latter may be less straightforward for translators when making post-editing decisions, since a threshold that sets a point from which post-editing becomes translating from scratch \citep{do2020differentiating} has yet to be widely established.

The possible advantages of adopting QE to facilitate the MTPE workflow extend beyond streamlining the initial assessment of MTPE’s cost-effectiveness. By providing information about the estimated quality of MT outputs, QE may help translators to allocate their efforts more effectively and focus on the outputs that deserve editing. On the one hand, they can spend minimal time on the least likely problematic MT outputs that require little intervention, avoiding making preferential edits. On the other hand, they can avoid wasting time reviewing and attempting to fix bad MT outputs unsuitable for post-editing \citep{moorkens2015correlations,specia-etal-2009-estimating}. In addition, QE may free up time for translators to focus on tasks that are difficult to automate, such as creative translation. Translators themselves have also expressed the need for CAT tools to present MT quality information or to highlight problematic MT outputs requiring attention \cite{Moorkens13,Moorkens17,Vieira2018}. While QE is not yet fully accurate in the realistic scenarios, its performance has steadily improved in recent years, particularly at the sentence-level \cite{blain-etal-2023-findings,specia-etal-2020-findings-wmt,specia-etal-2021-findings,zerva-etal-2022-findings}. Furthermore, the development of neural metrics alongside large language models offers the potential to further improve the accuracy and usability of QE \citep{zerva-etal-2024-findings}.

Despite the potential benefits and advancements of QE, it seems that QE has yet to be widely integrated in the real post-editing settings \citep{gilbert2022}. One possible reason for this is the scarcity of CAT tools that can effectively incorporate QE. While some CAT tools, such as Trados, have recently started offering QE information, it is usually not freely accessible, posing a challenge to the widespread adoption of QE. Moreover, there is limited empirical evidence supporting the usefulness of QE in enhancing MTPE workflow, which makes it challenging for translators to embrace QE, as they may be uncertain about its practical value and impact on their work. In real-world applications, various factors could influence the effectiveness of QE, such as translators’ attitudes towards QE, the accuracy of QE information, and the actual quality of MT outputs. Therefore, a critical question arises: to what extent and under what conditions can QE facilitate the MTPE processes? 

In light of the above, this study investigates the usefulness of sentence-level QE\footnote {Based on the granularity of assessment, QE models can be classified into word, sentence, and document levels. This study specifically focuses on sentence-level QE.}  in the context of English-Chinese MTPE, taking both productivity and users’ perceptions into consideration. It is expected that the current research can provide a more detailed understanding of QE’s application in aiding post-editing tasks, shedding new lights on its strengths and limitations. This insight will contribute to a more effective integration of QE into the MTPE workflow, enhancing efficiency for translators. Specifically, it focuses on the following three research questions:
1. What is the impact of sentence-level QE on post-editing time? 
2. Is the impact of sentence-level QE on post-editing time consistent across different conditions, in particular, varying levels of MT quality and translation expertise?
3. What are users’ perceptions of the usefulness of sentence-level QE in post-editing?

\section{Related Work}

Existing research on the usefulness of sentence-level QE in the context of MTPE has primarily focused on its impact on MTPE productivity and translators’ perceptions. While studies suggest that QE has the potential to enhance productivity, the evidence remains limited and mixed. For instance, \citet{huang-etal-2014-adaptive} observed a 10\% productivity increase when QE information was provided during post-editing tasks. However, this improvement was measured against a human translation condition rather than a post-editing condition without QE. In other words, the productivity gains resulted from a combined effect of MT and QE, making it unclear how much QE alone directly contributed to the observed improvement. Similarly, \citet{turchi-etal-2015-mt} found a slight increase in post-editing speed, but this increase was not statistically significant. In \citeposs{informatics8030061} study, post-editing with QE resulted in lower average post-editing time, fewer keystrokes, and higher translation quality compared to the condition without QE. Despite these positive findings, the study did not report the statistical significance of these differences, which leaves the robustness of the improvements uncertain. \citet{lee-etal-2021-intellicat} explored QE within IntelliCAT, a CAT interface that provides three intelligent features, namely QE, translation suggestion, and word alignment. The results indicated a significant improvement in post-editing efficiency when working with IntelliCAT. However, as the tool incorporated both word-level and sentence-level QE alongside with other two features, it was difficult to determine the extent to which segment-level QE contributed to the increased post-editing speed. 

In addition to productivity, it is essential to explore how translators perceive the usefulness of QE and the challenges they encounter when interacting with it. While earlier surveys revealed translators’ interest in using QE information \citep{Moorkens13,Moorkens17,Vieira2018}, few studies have investigated the perceptions of translators after letting them actually work with QE during post-editing, and the findings have been inconclusive. \citet{Escartín17} collected translators’ opinions on QE and revealed generally negative attitudes towards its usefulness. However, the reasons behind these negative results were not examined in this study. By contrast, most of the participants in \citet{lee-etal-2021-intellicat} expressed positive views on QE, particularly regarding its usefulness for proofreading purposes, such as double-checking the potential translation errors.

Apart from investigating the general impact of QE, efforts have also been made to consider additional factors and examine whether the usefulness of QE varies under specific conditions. For instance, given that QE was found to contribute only slight and insignificant global productivity gains in \citeposs{turchi-etal-2015-mt} study, the authors conducted an additional analysis to explore whether these marginal gains might become more pronounced under certain conditions. The analysis incorporated the length of source text (ST) and the quality of MT outputs, and the results suggested that QE led to significant productivity gains when the sentences were of medium length and had HTER\footnote{HTER (Human-targeted Translation Edit Rate) is a widely-used metric for assessing MT quality, which quantifies the number of edits necessary to transform the MT output into a good translation \citep{snover-etal-2006-study}. It ranges from 0 to 1, with lower HTER representing higher MT quality.} values between 0.2 and 0.5. The accuracy of QE has also been examined, with somewhat conflicting results. \citet{Escartín17} found that QE, especially good QE that provided a predicted quality score close to the actual score, significantly decreased post-editing time. However, \citet{teixeira-obrien-2017-impact} reported that no significant effect was introduced by QE, even when it was accurate. In addition, while not explicitly addressed as a variable of interest, \citet{informatics8030061} presented data pertaining to translation experience. In this study, despite varying levels of experience, all translators, with only one exception, increased their post-editing speed when QE information was provided.

In conclusion, there is a notable lack of empirical research on the effectiveness of presenting sentence-level QE information within the MTPE context, and the findings to date have been inconsistent. While considering additional factors has provided a more nuanced understanding of QE’s impact on post-editing efficiency, more research is warranted. It should be noted that most previous studies have relied on basic statistical analyses, which may not fully capture the true impact of QE. Additionally, while professional translators have been the focus of these studies, the way student translators utilise QE information in the MTPE workflow has yet to be investigated, which can provide valuable insights into translation education.

\section{Research Design and Methodology}
\subsection{Participants}
Thirty-one first-year Master in Translation and Interpreting (MTI) students (6 males, 25 females) participated in the post-editing experiments. The average age of the participants was 23 years (range = 21-33, SD = 2.4). All students used Chinese as their L1 and English as L2, and have passed the Test for English Majors at Band4 (TEM4). Although they were in the same year of study, their translation expertise varied, as reflected by the levels of the China Accreditation Test for Translators and Interpreters (CATTI) \footnote {Recipients of CATTI Level 3 (translator) certificate are expected to complete general translation work, while those with a Level 2 certificate should be capable of handling complex translation tasks within a particular domain (\url{http://www.catticenter.com/cattiksjj/1848})}they had achieved. To be more specific, 23 participants had passed the CATTI Level 3 (Translator), while 8 had passed the CATTI Level 2 (Translator). However, none of them had worked as professional translators. While the participants had limited experience with MTPE, they generally held a positive attitude towards it, with an average rating of 6.16 (SD=0.86) on a seven-point scale, where ‘7’ indicated a very positive attitude.

\begin{table*}
  \centering
  \resizebox{\textwidth}{!}{
  \begin{tabular}{cccccc}
    \hline
     &  \multicolumn{4}{c}{Source Text}& MT Output\\ 
      \hline
       &  Word Count&  Average Sentence Length&  FRE&  CAREC& MT Quality (mean/sd)\\ 
    \hline
    \textbf{Task1 (without QE)}&  \textbf{304.00}
&  \textbf{13.62}&  \textbf{63.32}&  \textbf{0.14}
& \textbf{2.58/0.10}
\\ 
 Text1&  48.00
&  12.00&  80.09&  0.13
& 2.60/0.20
\\ 
           Text2&  58.00
&  19.33&  44.27&  0.24
& 2.44/0.20
\\ 
           Text3&  83.00
&  10.38&  63.80&  0.01
& 2.67/0.00
\\ 
           Text4&  115.00
&  12.78&  65.13&  0.19
& 2.60/0.10
\\ 
    \hline
           \textbf{Task2 (with QE)}&  \textbf{295.00}
&  \textbf{12.90}&  \textbf{59.20}&  \textbf{0.11}
& \textbf{2.56/0.19}
\\ 
           Text5&  57.00
&  11.40&  58.64&  0.11
& 2.33/0.34
\\ 
           Text6&  48.00
&  16.00&  61.93&  0.16
& 2.67/0.14
\\ 
 Text7& 78.00
& 13.00& 70.32& -0.01
&2.75/0.13
\\ 
 Text8& 112.00
& 11.20& 45.91& 0.19
&2.48/0.07
\\ 
    \hline
  \end{tabular}}
    \caption{Summary of ST complexity and MT quality of the materials (a higher FRE score suggests lower complexity, while a higher CAREC score implies higher complexity) }
  \label{tab:accents}
\end{table*}

\subsection{Materials}
Given that this study focuses on the impact of QE on MTPE, it is essential to ensure the comparability between the materials used for the MTPE task without QE (Task 1) and the task with QE (Task 2). Specifically, textual characteristics, including ST complexity and MT quality, were controlled at a similar level across tasks, as suggested by previous research \citep{liu2024,jia2022interaction}. Each task \footnote {The materials, data, and script of statistical analyses used in the study are available at \url{https://github.com/jam0127/QEresearch.}} consisted of four short, self-contained news texts that required no specialist knowledge for post-editing. As shown in Table~\ref{tab:accents},  word count, average sentence length, and readability scores indicate that the two tasks were comparable in terms of ST complexity. Text readability was measured using two formulas: the Flesch Reading Ease (FRE) formula \citep{rudolph}, a traditional readability formula, and the Crowdsourced Algorithm of Reading Comprehension (CAREC) \citep{crossley2019moving}, a newer formula. These two metrics focus on different aspects of text complexity: FRE relies on word length and sentence length, while CAREC is based on features pertaining to lexical sophistication and text cohesion. Therefore, the readability scores may vary when measured by different formulas. For this study, Task 1 received a higher FRE score than Task 2 on average, suggesting slightly lower complexity. However, according to CAREC, Task 1 was judged to be slightly harder to read. Despite these minor variations, the readability scores across both tasks were similar overall. Therefore, we concluded that the tasks were comparable in terms of readability.

The STs were translated by Baidu Translate, a mainstream NMT engine. Three second-year MA students in translation participated in the MT quality evaluation. They were not involved in the post-editing experiments. All of them had prior experience in annotating MT errors and had passed the CATTI Level 2 (Translator). The MT outputs were rated at the segment level using a three-point scale: a score of ‘1’ suggested that the outputs require extensive editing or complete re-translation, while a score of ‘3’ indicated minimal or no editing was needed. The inter-rater agreement was strong and significant (Kendall’s W=0.705, p<0.05). Table~\ref{tab:accents} shows that the overall MT quality was comparable between the two tasks. 

\subsection{Research Procedures}
\begin{figure*}
    \centering
    \includegraphics[width=1\linewidth]{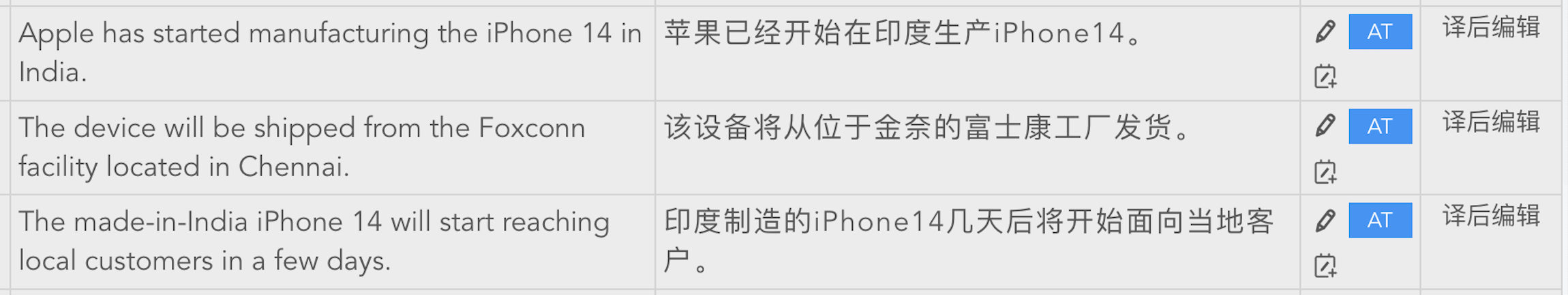}
    \caption{The YiCAT interface (Task 1, without QE information)}
    \label{fig:enter-label1}
\end{figure*}

\begin{figure*}
    \centering
    \includegraphics[width=1\linewidth]{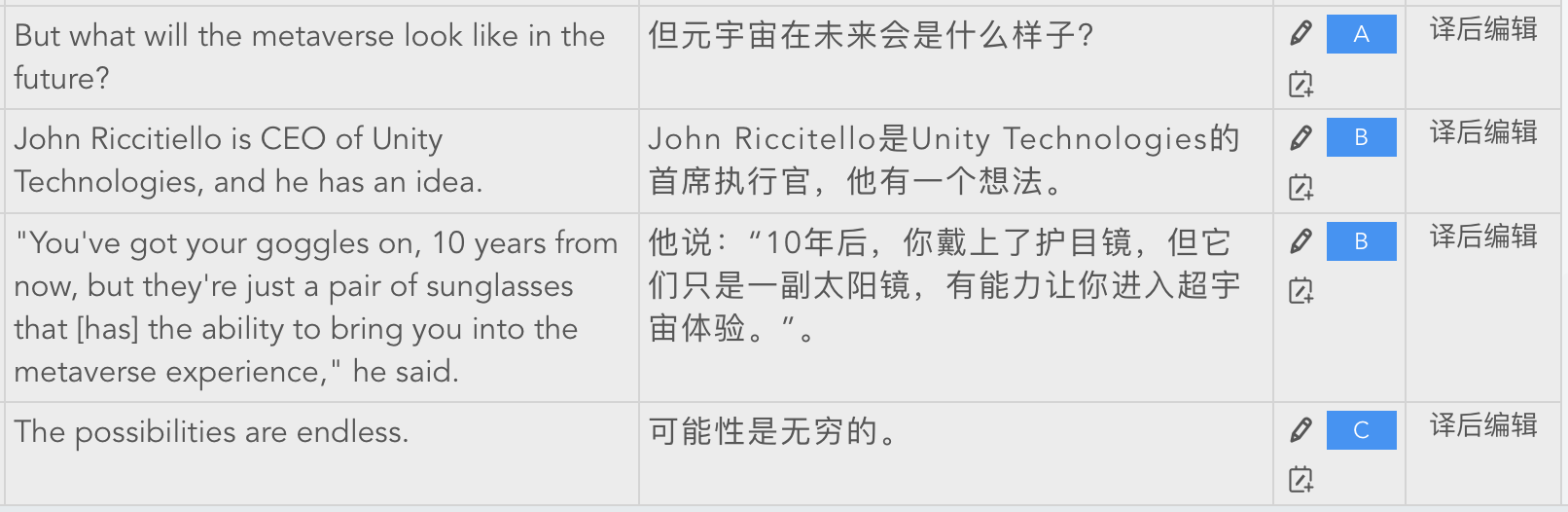}
    \caption{The YiCAT interface (Task 2, with QE information)}
    \label{fig:enter-label2}
\end{figure*}

To ensure the ecological validity of this study, we adopted YiCAT, a Chinese online CAT platform employed in the realistic translation scenario, along with its QE system. Since 2022, YiCAT has integrated QE as an optional feature within its interface, allowing translators to choose whether to display the information of estimated MT quality. Figure~\ref{fig:enter-label1} and Figure~\ref{fig:enter-label2} illustrate the interface used by participants for Task 1 and Task 2 respectively. The only difference between the two task interfaces lies in the third column (from left to right). In Task 1, it did not display any QE information (AT in this column is short for automatic translation). In Task 2, the column presented QE scores: “A” indicated that MT output is of good quality and requires minimal editing, “B” denoted medium-quality MT requiring moderate editing, and “C” represented poor-quality MT outputs that need extensive editing or retranslation. All editing actions were performed in the target text area (the second column from left to right).

The post-editing experiment was conducted on the campus of Guangdong University of Foreign Studies in October 2022. One day prior to the experiment, participants received a video tutorial on using YiCAT \footnote{It is important to note that in YiCAT, the time spent on a segment would not be recorded if no edits were made to the segment. To ensure that the post-editing time for each segment was captured, participants were instructed to type ‘1’ at the end of a segment if they believed the MT output required no editing. This additional step was also emphasised during the training session conducted before each post-editing experiment.} and were required to complete a practice task to familiarise themselves with the platform. On the day of the experiment, a short guide on MTPE was first introduced to the participants, which covered key topics such as the concept of MTPE, differences between light and full MTPE, MTPE guidelines, MT quality assessment, and QE. Most importantly, participants were explicitly informed about the meaning of QE scores and encouraged to use them critically, since they may not always be accurate. This explanation was provided before both Task 1 and Task 2 to ensure a clear understanding of QE, even though QE information was only available in Task 2.

Participants were required to perform full MTPE according to GB/T 40036-2021: Translation services — Post-editing of machine translation output -Requirements\footnote{\url{https://www.gbstandards.org/China_standard_english.asp?code=GB/T\%2040036-2021\&id=49840}}, the Chinese national standard for post-editing. Then, a warm-up task was conducted by participants, followed by Task 1. A week later, similar procedures were followed for Task 2. Participants were again reminded of the MTPE guidelines, the interpretation of QE scores, and task requirements. Task 2 was conducted after a warm-up task. No time limits were imposed on the tasks, but participants were suggested to finish them as soon as possible. External resources, such as dictionaries, were prohibited. 

Within two days of completing Task 2, twelve volunteers participated in one-on-one interviews. Participants were encouraged to share their experiences and perceptions of QE freely. The interviews followed a semi-structured outline, covering questions such as “when do you typically check QE scores (e.g., before reading the ST and MT; after reading the ST and MT; or after editing the MT)?”; “to what extent do you trust and rely on QE scores?”; and “do you think that adopting QE in post-editing tasks can increase efficiency?”. 

\subsection{Data Processing and Statistical Analysis}
The data analysis was conducted at the segment level using the statistical software R \citep{R}. Linear Mixed Effects Regression (LMER) models were employed to investigate the impact of QE on post-editing time. To address the first research question, a LMER model was built with task type (Task 1: without the aid of QE; Task 2: with the aid of QE) as the fixed effect. For the second research question, two additional LMER models were built. The first one included task type, MT quality, and their interaction as fixed effects. The second model included task type, translation expertise (students with CATTI Level 2 were classified as having higher expertise, while those with CATTI Level 3 were considered as having lower expertise), and their interaction as fixed effects. All models used post-editing time as dependent variable, with participants and segments as random effects. Prior to model fitting, post-editing time was normalized by the number of words in the ST and transformed to approximate a normal distribution. Subsequently, the models were constructed, and their residuals were checked for normality and homoscedasticity.

It is important to note that, since there was only one segment being rated as low-quality by human raters, data pertaining to this segment was excluded from the model that included task type, MT quality, and their interaction as fixed effects. Therefore, the analysis of the interaction effect between MT quality and QE is limited to the cases of medium- and high-quality MT.

The interview data was transcribed and coded according to the outlined questions, serving as a complementary source to the post-editing experiment data in the current study. Due to the particular research focus and effort constraints, the analysis focused on participants’ perceptions regarding the potential of QE to increase MTPE efficiency.

\section{Results and Discussion}

\subsection{The Impact of QE on Post-editing Time}
\begin{figure}
    \centering
    \includegraphics[width=\columnwidth]{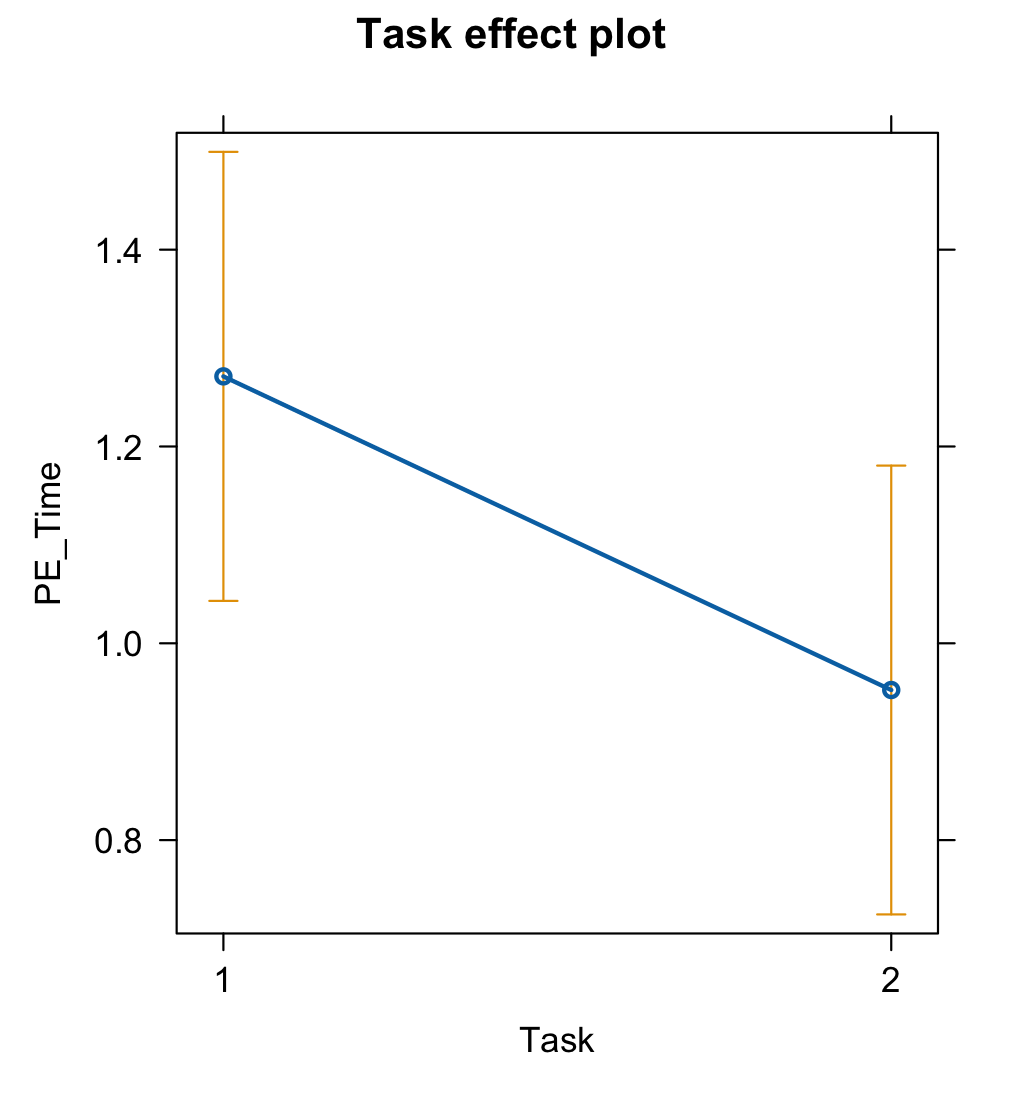}
    \caption{The effect of task type (Task 1: without QE, Task 2: with QE) on post-editing time }
    \label{fig:enter-label3}
\end{figure}

In order to assess the overall impact of QE, we first analysed the LMER model with task type as the main effect. As shown in Figure~\ref{fig:enter-label3}, Task 2 took less time than Task 1. To be more specific, the average time was 0.95s per word (SD=0.94) for Task 2, while it was 1.27s (SD=1) for Task 1. The main effect of the model was statistically significant (t=-2.34, p=0.02<0.05), suggesting that the use of QE information reduced post-editing time. 

This reduction in post-editing time indicates the practical utility of QE information in enhancing translation efficiency. As mentioned previously, the significant impact of QE can be attributed to its potential to save translators time in evaluating MT quality and deciding whether to post-edit the outputs or discard them and translate from scratch. Additionally, QE may assist translators in quickly identifying and revising potentially erroneous segments, thereby prioritising and streamlining error correction. These findings align with previous research \citep{huang-etal-2014-adaptive,lee-etal-2021-intellicat,specia-2011-exploiting}, which emphasised the role of QE in reducing processing time and enhancing workflow efficiency in post-editing tasks.

\subsection{The Interaction Effect between QE and MT Quality}
Having preliminarily established the significant impact of QE on post-editing time, we were interested in whether this effect remains consistent across different conditions. To address this question, we considered one important factor that could potentially influence post-editing time: the quality of MT outputs. 

\begin{figure}
    \centering
    \includegraphics[width=\columnwidth]{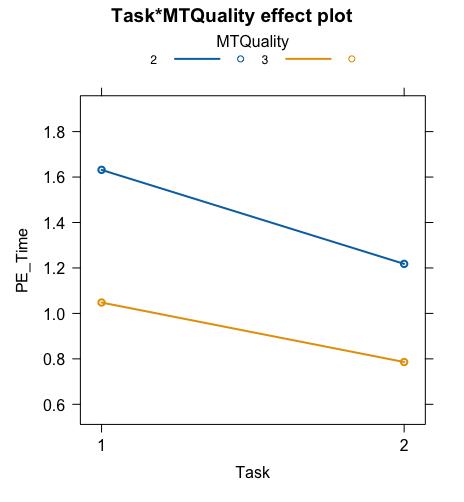}
    \caption{The interaction effect between MT quality (MTQuality=2: medium-quality MT, MTQuality=3: high-quality MT) and task type (Task 1: without QE, Task 2: with QE) on post-editing time}
    \label{fig:enter-label4}
\end{figure}

As illustrated in Figure~\ref{fig:enter-label4} and supported by the model results, the interaction effect between MT quality and task type was not significant (t=0.62, p=0.54>0.05), indicating that the impact of task type on post-editing time remained consistent regardless of the MT quality levels. The effect of task type was significant (t= -2.13, p=0.04<0.05), with participants spending less time on Task 2 than on Task 1. MT quality also had a significant impact on post-editing time (t=-3.45, p=0.001<0.01). Specifically, MT outputs with lower quality led to longer post-editing time, which aligns with the findings of \citet{gaspari-etal-2014-perception}, \citet{o2011towards}, and \citet{tatsumi-2009-correlation}. 

The results suggest that post-editing with QE is consistently and significantly faster than without QE, no matter if the MT outputs are of medium or high quality. Our findings are partially consistent with those of \citet{turchi-etal-2015-mt}, who observed that QE significantly increased post-editing speed when the HTER value was between 0.2 and 0.5. Although \citet{turchi-etal-2015-mt} did not categorise MT quality into high, medium, and low, they adopted a binary classification with a threshold of 0.4 to distinguish between editable and useless MT. In our study, both high- and medium-quality were considered “editable”. Therefore, the 0.2 to 0.5 range identified by \citet{turchi-etal-2015-mt} overlaps to some extent with the quality levels examined in the current model. 

These consistent efficiency gains suggest that QE can offer practical advantages across various scenarios. For instance, when dealing with high-quality MT outputs, presenting QE information may prevent translators from making unnecessary preferential edits. Such edits require certain effort and time but do not lead to increased translation quality and can sometimes even be detrimental \citep{koponen2019product}. Therefore, if translators know in advance that the segment they are working with is of high-quality, they are more likely to spend less time on it, thereby increasing post-editing efficiency. In the case of medium-quality MT outputs, QE can potentially help translators allocate their attention more effectively by identifying segments that are worthy of intervention. This allows them to concentrate on the task of editing itself, rather than second-guessing the overall quality of MT. Such a targeted approach can streamline the MTPE process, enabling translators to work more efficiently and effectively.

\subsection{The Interaction Effect between QE and Translation Expertise}
\begin{figure}
    \centering
    \includegraphics[width=\columnwidth]{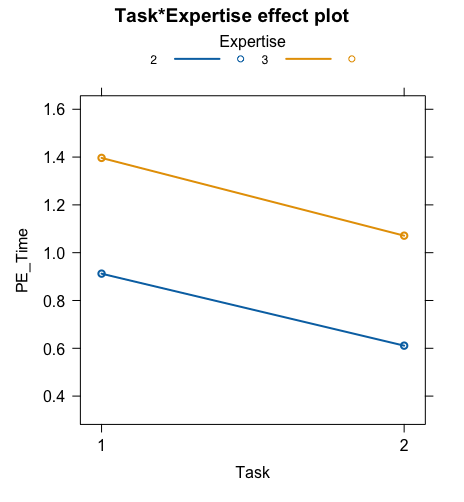}
    \caption{The interaction effect between translation expertise (Expertise=2: students with CATTI Level 2, Expertise=3: students with CATTI Level 3) and task type (Task 1: without QE, Task 2: with QE) on post-editing time }
    \label{fig:enter-label5}
\end{figure}

In addition to examining the quality of MT outputs, we also investigated the role of translation expertise in influencing the effectiveness of QE on post-editing time. The results indicated that the interaction effect between translation expertise and task type was not significant (t=-0.26, p=0.80>0.05). In other words, the impact of task type did not differ across student translators with varying levels of expertise. As shown in Figure~\ref{fig:enter-label5}, Task 2 required less time than Task 1 in both groups, suggesting that QE may have contributed to reduced post-editing time. The model results further indicated a marginally significant effect of task type (t= -1.97, p=0.05<0.1), pointing to a potential trend toward greater efficiency when QE information was available. In addition, translation expertise had a significant impact on post-editing time (t=3.46, p=0.001<0.01), with students with a higher level of expertise completing tasks more quickly than those with less expertise.

The results indicate that translation students, irrespective of their expertise levels, may have experienced similar improvements in speed from the presence of QE information, although the observed advantages were only marginally significant. This finding aligns with \citeposs{informatics8030061} study, where nearly all professional translators across varying experience levels completed post-editing tasks more quickly with the aid of QE, except for one translator who maintained the same speed regardless of QE availability. One possible explanation is that QE provides explicit cues, so the cognitive processes involved in interpreting and utilising QE information may be straightforward, thus not necessitating advanced translation expertise. Moreover, participants in this study were informed that QE is not infallible and can make mistakes, and they were asked to engage with the information critically. It is therefore plausible that the students followed the instructions and integrated the QE information effectively, leading to productivity gains across the board. However, these findings warrant further validation, particularly through comparisons between professional and student translators, to confirm their generalisability.  

\subsection{Users’ perceptions}
This section focuses on participants' views on the potential of QE to increase MTPE efficiency. As summarised in Table~\ref{tab:my_label}, the interview data reveal a range of opinions, including some conflicting perspectives. Specifically, 66.7\% (8) of the interviewees believed that QE could improve MTPE quality. Interestingly, while much of the previous literature has focused on QE’s impact during the pre-processing stage (i.e. the process of evaluating MTPE’s cost-effectiveness), participants in this study highlighted potential applications of QE in the later stages of MTPE. For example, interviewees reported using QE to check whether they had overlooked any MT errors, which is consistent with the results of \citet{lee-etal-2021-intellicat}. Additionally, one participant used QE to validate her evaluation of MT quality, noting that this validation increased her confidence in the decisions regarding whether to edit MT outputs or not. However, among these eight interviewees, perceptions of QE's impact on MTPE speed were divided: half felt it helped them work faster, while the other half did not notice any meaningful improvement.

\begin{table*}
    \centering
    \resizebox{\textwidth}{!}{
    \begin{tabular}{ccc} 
     \hline
         \textbf{Views}& \textbf{N} & \textbf{Main Reasons}\\ 
          \hline
         Increasing quality but not necessarily the speed&  4(33.3\%)& Validating translators’ own evaluation of MT quality; assisting quality check\\ 
         Increasing both quality and speed&  4(33.3\%)& Saving time and effort for more difficult translations; assisting quality check\\ 
         No impact&  2(16.7\%)& A firm belief in translators’ own evaluation of MT quality\\ 
         Increasing speed&  1(8.3\%)& The ability of QE to highlight potentially erroneous MT\\ 
         Decreasing speed&  1(8.3\%)& Low accuracy of QE; distrust of QE\\ 
    \hline
    \end{tabular}}
    \caption{Users’ perceptions of QE’s potential in increasing MTPE efficiency}
    \label{tab:my_label}
\end{table*}

Notably, two participants perceived that QE had no impact on their MTPE processes, as they were very confident in their own assessment of MT quality. Finally, two students commented solely on QE’s impact on speed without referencing its effect on quality. Their views were contradictory: one believed QE increased speed by highlighting potentially erroneous MT segments, while the other felt that QE slowed her down, citing distrust in its accuracy and a belief that the tool produced unreliable assessments. Although previous studies have not demonstrated that inaccurate QE negatively affects post-editing efficiency \citep{Escartín17,teixeira-obrien-2017-impact}, particularly in comparison to working without QE, the interview data from this study suggest that poor QE accuracy may adversely impact users’ experience and even reduce post-editing speed.

\section{Conclusions and Future Work}
Motivated by the potential benefits of QE in streamlining MTPE workflow, the current study preliminarily explored the usefulness of sentence-level QE in increasing post-editing speed and gathered student translators’ views about its application. Three major findings emerged. First, QE significantly reduced post-editing time, and no significant interaction effects were found between QE and MT quality or between QE and translation expertise. In other words, the impact of QE remained consistent across MT outputs of medium and high quality and among students with varying levels of translation expertise. This stability implies that the advantages of QE in reducing post-editing time are likely to be broadly applicable. Second, the benefits of using QE in post-editing extend beyond highlighting problematic MT segments, it can also validate translators’ own evaluations of MT quality and assist in quality checking. These findings shed new light on how translators can integrate QE information into the MTPE workflow to enhance overall efficiency. Finally, although this study did not explicitly examine the impact of QE's accuracy, interview data indicate a potential detrimental effect of inaccurate QE on post-editing processes. However, this finding requires further empirical validation. 

This study has several limitations that open avenues for future research, particularly regarding the number and diversity of texts and participants, the range of factors considered, and the indicators used to measure MTPE efficiency. In future research, we aim to expand the sample size by including more participants and a wider variety of text types to better understand the conditions under which QE proves most beneficial. Comparisons between student and professional translators will also be conducted to assess whether the benefits of QE differ when larger differences in expertise are present. Furthermore, low-quality MT outputs will be included to examine whether QE can still enhance post-editing efficiency in such cases. Other factors, such as the accuracy of QE and the score levels assigned by QE, will also be considered. Additionally, eye-tracking data, which can capture how translators allocate their cognitive resources when presented with QE information, will be collected to gain a more detailed understanding of QE’s impact on the post-editing processes.

In conclusion, this study provides preliminary evidence for the usefulness of sentence-level QE in the MTPE context. Instead of simply saying “no” to QE, we should embrace its potential and investigate how to optimise its integration into MTPE workflows. As one interviewee aptly remarked, \textit{“If we have access to such information, why not use QE?”} This perspective encapsulates the pragmatic value of QE and underscores the need for further exploration into its role in enhancing MTPE efficiency.

\section*{Acknowledgments}
The work described in this paper was partially supported by the National Social Science Fund of China (“A Study on Quality Improvement of Neural Machine Translation”, Grant reference: 22BYY042) and a grant from CBS Departmental Earnings Project of the Hong Kong Polytechnic University (Project title: Predicting Machine Translation Post-Editing Effort with Source Text Characteristics and Machine Translation Quality: An Eye-Tracking and Key-Logging Study; Project No.: P0051091).

% Bibliography entries for the entire Anthology, followed by custom (mtsummit25) entries
%\bibliography{anthology,mtsummit25}
% Custom bibliography entries only
\bibliography{mtsummit25}

\end{document}